\documentclass{article}
\usepackage{ijcai25}
\usepackage{xcolor}
\usepackage{graphicx}
\usepackage{colortbl}
\usepackage{adjustbox}
\usepackage{arydshln}
\usepackage{times}
\usepackage{booktabs}
\usepackage{multirow}
\usepackage{latexsym}
\usepackage{amssymb} 
\usepackage{times}
\usepackage{soul}
\usepackage{url}
\usepackage[hidelinks]{hyperref}
\usepackage[utf8]{inputenc}
\usepackage[small]{caption}
\usepackage{amsmath}
\usepackage{amsthm}
\usepackage{booktabs}
\usepackage{algorithm}
\usepackage{algorithmic}
\usepackage[switch]{lineno}
\usepackage{microtype}

\title{Modality-Fair Preference Optimization for Trustworthy MLLM Alignment}

\author{
Songtao Jiang$^1$ \and
Yan Zhang$^2$ \and
Ruizhe Chen$^1$ \and
Tianxiang Hu$^1$ \\
Yeying Jin$^3$ \and
Qinglin He$^1$ \and
Yang Feng$^4$ \and
Jian Wu$^1$ \and 
Zuozhu Liu$^{1\dagger}$ \\
\affiliations
$^1$Zhejiang University \\
$^2$ByteDance\\
$^3$National University of Singapore \\
$^4$Angelalign Inc., China\\
\emails
\{songtao.22, zuozhuliu\}@intl.zju.edu.cn
}

\begin{document}

\maketitle
\let\thefootnote\relax\footnote{† Corresponding author: zuozhuliu@intl.zju.edu.cn}

\begin{abstract}
Multimodal large language models (MLLMs) have achieved remarkable success across various tasks. However, separate training of visual and textual encoders often results in a misalignment of the modality. Such misalignment may lead models to generate content that is absent from the input image, a phenomenon referred to as hallucination. These inaccuracies severely undermine the trustworthiness of MLLMs in real-world applications. Despite attempts to optimize text preferences to mitigate this issue, our initial investigation indicates that the trustworthiness of MLLMs remains inadequate.
Specifically, these models tend to provide preferred answers even when the input image is heavily distorted. Analysis of visual token attention also indicates that the model focuses primarily on the surrounding context rather than the key object referenced in the question. These findings highlight a misalignment between the modalities, where answers inadequately leverage input images. Motivated by our findings, we propose Modality-Fair Preference Optimization (MFPO), which comprises three components: the construction of a multimodal preference dataset in which dispreferred images differ from originals solely in key regions; an image reward loss function encouraging the model to generate answers better aligned with the input images; and an easy-to-hard iterative alignment strategy to stabilize joint modality training. Extensive experiments on three trustworthiness benchmarks demonstrate that MFPO significantly enhances the trustworthiness of MLLMs. In particular, it enables the 7B models to attain trustworthiness levels on par with, or even surpass, those of the 13B, 34B, and larger models. 
\end{abstract}

\section{Introduction}
Recent advances in multimodal large language models (MLLMs)~\cite{liu2024visual,wang2023cogvlm,jiang2024med} have achieved remarkable performance on diverse multimodal tasks. However, separate training of visual and textual encoders often causes a misalignment of the modality~\cite{cui2023holistic,guan2024hallusionbench} of MLLMs, which can cause the model to generate content that does not exist in the visual input—a phenomenon known as hallucination~\cite{bai2024hallucination}. This issue compromises the reliability of MLLMs in real-world applications~\cite{shah2019cycle}.

To address this issue, recent work has applied preference optimization to improve the alignment of modalities in MLLM~\cite{rafailov2024direct,naveed2023comprehensive}. These methods typically train models to favor accurate responses over those that contain erroneous image descriptions. They achieve this by using inaccurate outputs as rejected options to enhance learning from correct responses, thus improving the alignment of modality~\cite{wang2024mdpo,zhang2024automated}. 

\begin{figure}[t!]
    \centering
    \includegraphics[width=1\linewidth]{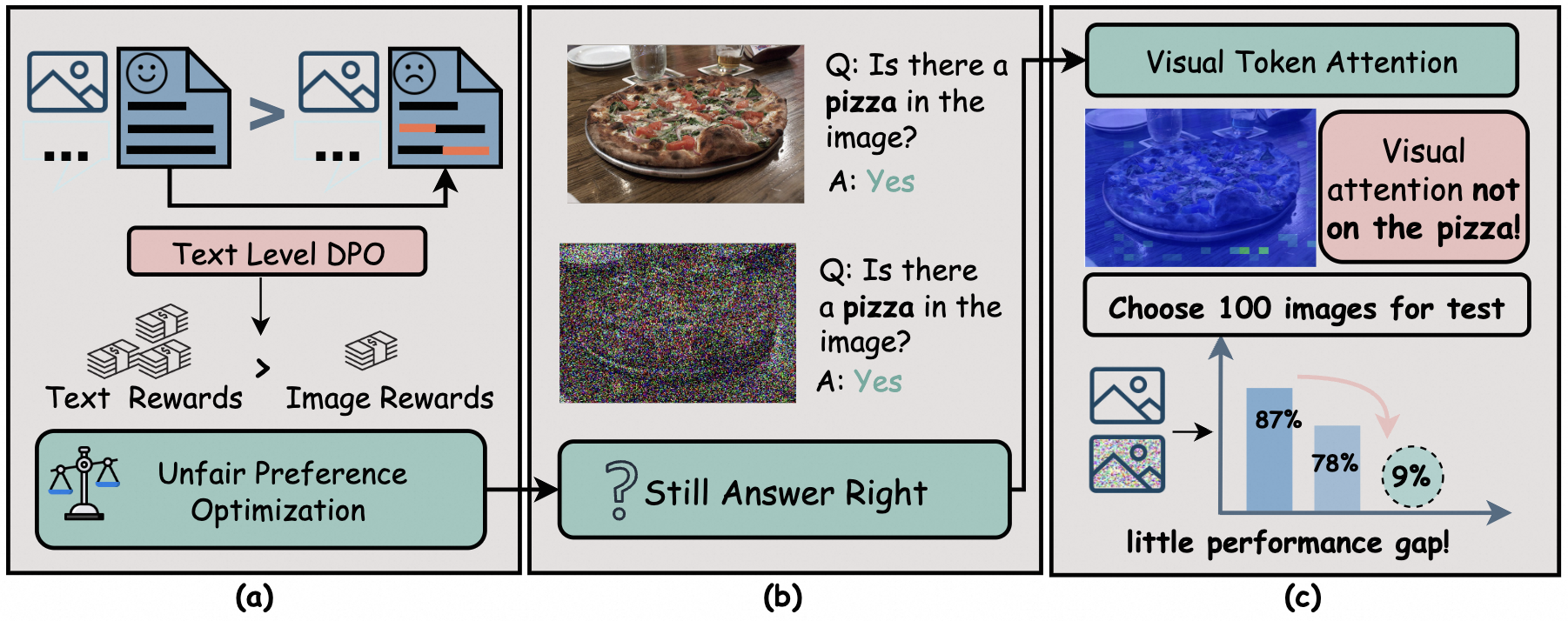}
\caption{(a) Current preference optimization (PO) in LLaVA-v1.5 results in imbalanced modality rewards. (b) After PO, LLaVA-v1.5 exhibits a persistent bias towards preferred responses, even with distorted image inputs. (c) Visual token attention analysis, supported by experimental evidence, indicates that MLLMs tend to memorize text preferences rather than achieving modality alignment.}
    \label{fig:intro_case}
\end{figure}

Although current preference optimization methods show success on trustworthiness benchmarks, they mainly rely on text preference optimization. This raises a critical question: \textit{What does the model learn during preference optimization when only text preferences are used? Is it merely memorizing text preferences or is it learning to align image and text data?} Our experiments shown in Figure~\ref{fig:intro_case} reveal that even with severe noise-degrading image content~\cite{leng2024mitigating}, the model consistently provides accurate answers. Across a sample of 100 images, the disparity in performance between the original and degraded images is only 9\%. Furthermore, the visualization of visual token attention shows that the model mainly focuses on the surrounding contexts rather than the key object mentioned in the question. These findings suggest that the model seems to lean more towards memorizing text preferences rather than achieving modality alignment after preference optimization. 

\begin{figure}[t!]
\centering
\includegraphics[width=1\linewidth]{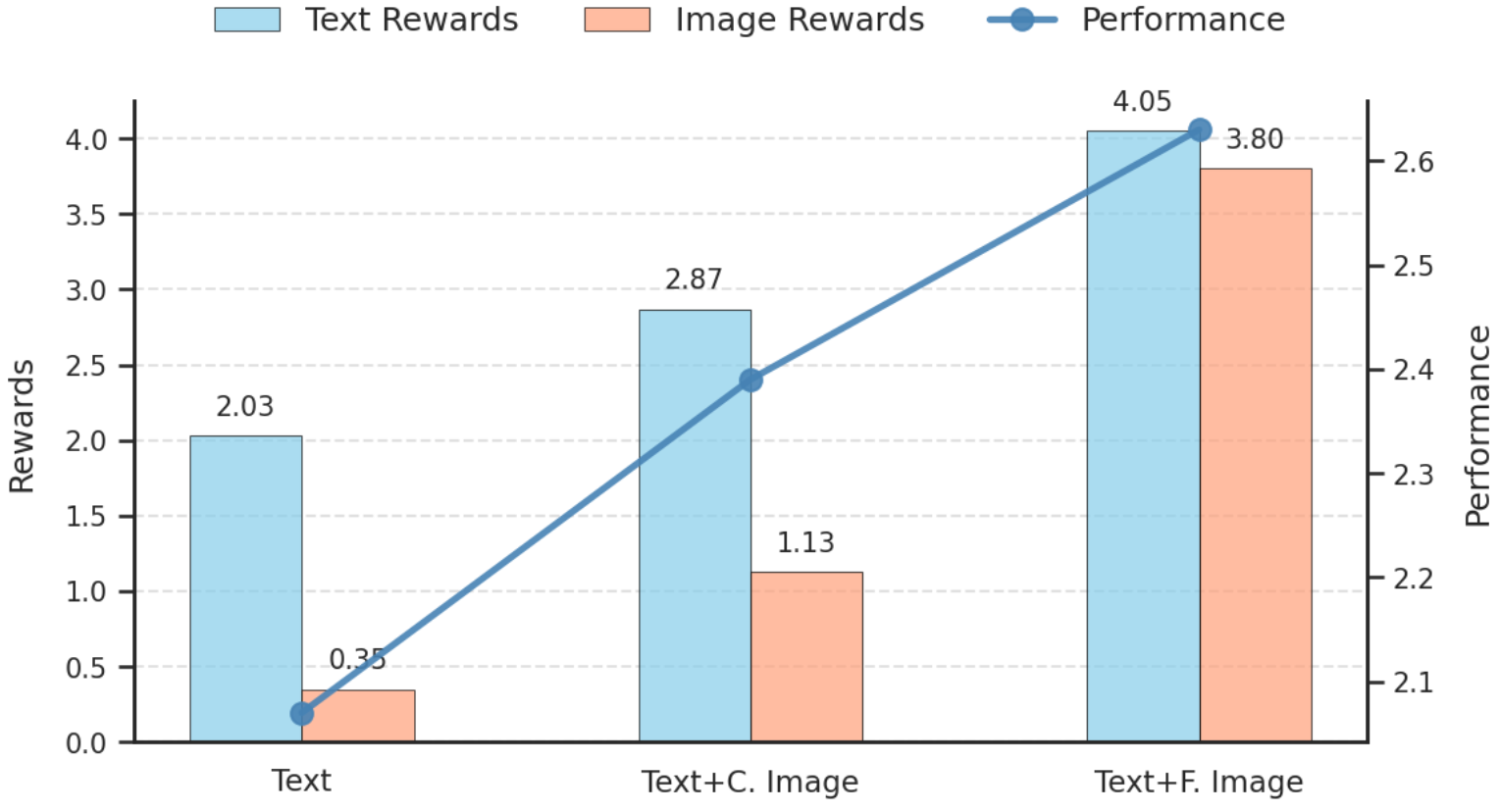}
\caption{"Text" refers to methods utilizing manually annotated text preference datasets (RLHF-V~\protect\cite{yu2024rlhf}). "Text + C. Image" involves image preference datasets generated by randomly cropping images~\protect\cite{wang2024mdpo}, while "Text + F. Image" uses fine-grained image preference datasets constructed with SAM and manual annotations. As the quality of image preference datasets improves, rewards across modalities become more balanced, leading to higher overall rewards and enhanced trustworthiness.}
\label{fig:intro_case2}
\end{figure}

We explore implementing more modality balanced preference optimization to encourage MLLMs to focus more on modality alignment for improved trustworthiness. As shown in Figure~\ref{fig:intro_case2}, we evaluated adding image-related rewards with a simple image preference dataset and a more fine-grained image preference dataset. As we progressively update the approach, the image and text rewards become higher and also more balanced, and the model's trustworthy performance consistently improves. The limited impact of simple image degradation may be due to the model quickly learning to distinguish these variations after only a few training steps, which leads it to continue prioritizing text preference optimization. These findings illustrate the effectiveness of more balanced optimization in achieving trustworthiness.
Motivated by our investigation, we propose Modality-Fair Preference Optimization (MFPO) for trustworthy MLLM alignment. To circumvent mere memorization of text responses, we propose to jointly optimize over both text and image preference in a modality-fair manner. In particular, we first construct a multimodal preference dataset encompassing both text and image preference data. Beginning with a text preference dataset, we characterize the relationship between words using a multipartite graph~\cite{jin2014complete,boudin2018unsupervised} to identify the top-$K$ keywords. These keywords are then linked to relevant image regions using a modified Segment Anything Model (SAM)~\cite{kirillov2023segment}. By introducing diffusion noise, we generate dispreferred image data that remains largely faithful to the originals while introducing meaningful differences~\cite{jin2014complete,dawande2001bipartite}, which are critical for preference optimization. Afterwards, to encourage fair alignment over both modalities, we devise a novel loss function consisting of text preference alignment, image preference alignment and a margin loss for stablized training. Furthermore, to tackle the instability in joint modality optimization, we propose an iterative alignment approach ~\cite{paas2003cognitive}. By categorizing training data into varying complexities using semantic entropy, we initially train the model on simpler data and gradually advance to more challenging samples.

Extensive experiments demonstrate the effectiveness of MFPO in improving model trustworthiness. Remarkably, MFPO empowers 7B LLaVA-v1.5 to achieve trustworthiness levels that are comparable to, or even exceed, those of significantly larger 13B, 34B, and larger-scale models. Specifically, when applied to models such as LLaVA-v1.5-13B, MFPO outperforms GPT-4V, achieving a substantial 40\% improvement on Object HalBench and setting new state-of-the-art results on both Object HalBench and AMBER benchmarks. Furthermore, LLaVA-v1.5-7B+MFPO and LLaVA-v1.5-13B+MFPO demonstrate superior performance over GPT-4V across five out of eight evaluation metrics. Comprehensive ablation studies further confirm the efficacy of the three core components in MFPO. Our work underscores the importance of balanced preference optimization over different modalities for MLLMs.

\begin{figure*}[h!]
\centering
\includegraphics[width=0.8\linewidth]{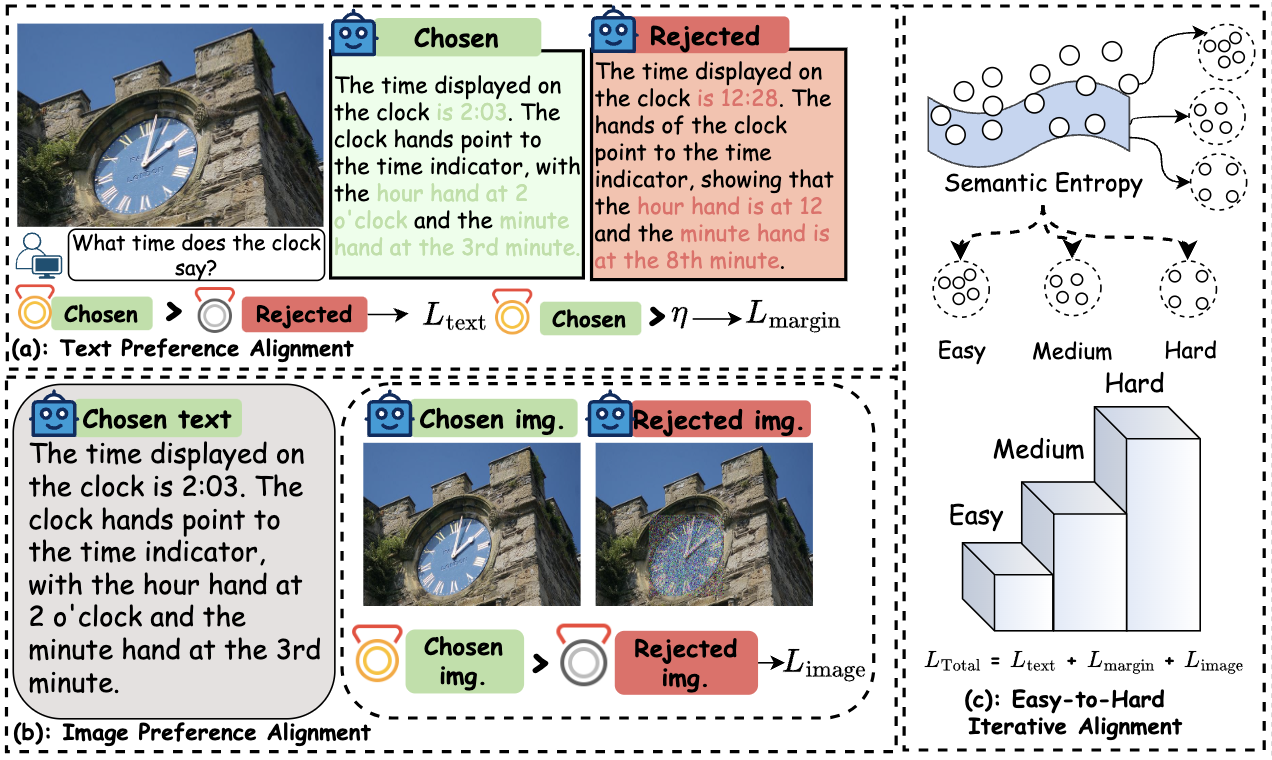}
\caption{Overview of the MFPO framework. (a) Chosen and rejected text responses are compared to compute text loss \( L_{\text{text}} \), with margin loss \( L_{\text{margin}} \) for reward stability. (b) Chosen and perturbed rejected images are compared, optimizing image preferences with loss \( L_{\text{image}} \). (c) Samples are categorized by difficulty using semantic entropy, facilitating progressive training. The total loss \( L_{\text{Total}} \) combines \( L_{\text{text}} \), \( L_{\text{margin}} \), and \( L_{\text{image}} \). }
\label{fig:model_pipe}
\end{figure*}
\section{Preliminaries}

\paragraph{Reinforcement Learning from Human Feedback (RLHF)} aligns models with human preferences using a reward model $r_{\phi}$~\cite{schulman2017proximal}, trained on pairwise preference data~\cite{christiano2017deep,ouyang2022training,casper2023open}. The model assigns higher rewards to preferred outputs, with the cross-entropy loss:
\begin{equation}
\label{RLHFRM}
    L_{\text{RM}} = -\log \left( \sigma \left( r_{\phi}(x, y_w) - r_{\phi}(x, y_l) \right) \right),
\end{equation}
where $\sigma(\cdot)$ is the logistic sigmoid, and $r_{\phi}(x, y_w)$ and $r_{\phi}(x, y_l)$ are the rewards for the preferred and less preferred outputs.

After training, the policy $\pi_\theta$ is optimized by maximizing the expected reward while regularizing it to prevent divergence from the reference policy $\pi_{\text{ref}}$, using a KL divergence penalty:
\begin{equation}
\resizebox{0.5\textwidth}{!}{$
    \max_{\pi_\theta} \mathbb{E}_{x \sim D, y \sim \pi_\theta(y|x)} \left[ r_{\phi}(x, y) - \beta D_{\text{KL}}(\pi_\theta(y|x) \parallel \pi_{\text{ref}}(y|x)) \right],
    $}
\end{equation}
where $\beta$ balances reward maximization and regularization. The KL divergence ensures policy stability~\cite{peng2023stabilizing}.

\paragraph{Direct Preference Optimization (DPO)} directly optimizes a policy \( \pi_\theta \) using preference data \( D \)~\cite{rafailov2024direct}, without relying on a reward model. 
Concretely, DPO derives a mapping between the reward \( r(x, y) \) and the policy \( \pi_\theta \):
\begin{equation}
\label{DPOr}
    r(x, y) = \beta \log \frac{\pi_\theta(y|x)}{\pi_{\text{ref}}(y|x)} + \beta \log Z(x),
\end{equation}
where \( Z(x) \) is the partition function, defined as $Z(x) = \sum_{y} \pi_{\text{ref}}(y|x) \exp\left(\frac{1}{\beta} r(x, y)\right) $.
Substitute Eq.~\ref{DPOr} into Eq.~\ref{RLHFRM}, we can derive the following loss:
\begin{equation}
\label{DPO_all}
\small
\begin{aligned}
L_{\text{DPO}}(\pi_\theta; \pi_{\text{sft}}) &= - \mathbb{E}_{(x, y_w, y_l) \sim D} \left[ \log \sigma \left( \beta \log \frac{\pi_\theta(y_w|x)}{\pi_{\text{sft}}(y_w|x)} \right. \right. \\
&\qquad\qquad\qquad - \left. \left. \beta \log \frac{\pi_\theta(y_l|x)}{\pi_{\text{sft}}(y_l|x)} \right) \right],
\end{aligned}
\end{equation}
where \( y_w \) and \( y_l \) denote the preferred and less preferred outputs, respectively. This approach bypasses the complexity of RLHF by allowing the policy gradient to be computed analytically, ensuring efficient alignment with human preferences~\cite{xu2024dpo,feng2024towards}.

The common approach in DPO for MLLMs~\cite{yu2024rlhf,yu2024rlaif,sun2023aligning,zhou2024aligning} involves concatenating the image \( m \) and question \( t \) into a single input \( x \), followed by optimization using Eq.~\ref{DPO_all}. Chosen responses \( y_w \) typically provide accurate, hallucination-free information, while rejected responses \( y_l \) often contain hallucinated details, making them less preferred.

\section{Modality-Fair Preference Optimization}
\subsection{Image Preference Data Generation}
\label{gen_da}
\label{sec:gen}
\begin{figure*}
    \centering
    \includegraphics[width=0.8\linewidth]{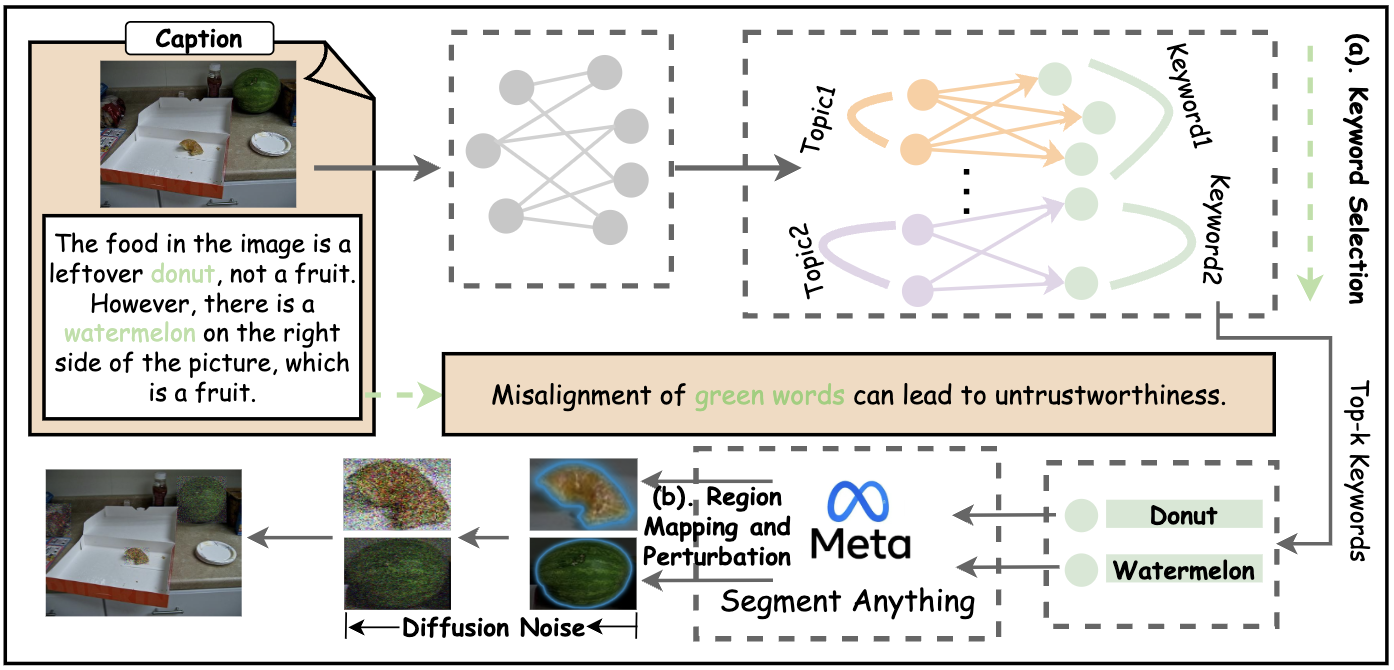}
    \caption{Our data generation pipeline consists of two stages: (a) Keyword Selection, where top-$K$ keywords are identified using a multipartite graph; and (b) Region Mapping and Perturbation, where selected regions are perturbed with diffusion noise to generate rejected images.}
    \label{fig:data_gen}
\end{figure*}
We first introduce the construction of our multimodal preference data as shown in Figure~\ref{fig:data_gen}. A key question is: \textit{what defines effective image preference data in DPO for MLLMs?}
Previous works on text preference data~\cite{chen2024noise} suggest that optimal preference data consists of chosen and rejected responses with moderate differences, as this balance enhances gradient effectiveness during training and helps the model capture subtle content distinctions. Inspired by this, our approach focuses on generating image preference data with fine-grained variation, which involves introducing perturbations only to the key regions of the original image most relevant to the chosen responses. The details of the hyperparameters in this section can be found in Supplementary Section 2.

\noindent{\textbf{Keyword Selection.}} To achieve balanced optimization of image and text preferences, we refine the construction of image preference data at the region level to better capture image preferences.
We first construct a multipartite graph~\cite{boudin2018unsupervised}, $G = (V, E)$, where nodes $V$ represent words and edges $E$ encode both positional and semantic relationships. The positional influence between two words $k_i$ and $k_j$ is defined as \( \theta_{ij} = \sum_{l_i \in \mathcal{L}(k_i)} \sum_{l_j \in \mathcal{L}(k_j)} \frac{1}{1 + |l_i - l_j|^\varphi} \), where \( \varphi \) controls the decay with distance. Semantic similarity is calculated using cosine similarity: \( S(k_i, k_j) = \frac{\mathbf{v}(k_i) \cdot \mathbf{v}(k_j)}{\|\mathbf{v}(k_i)\| \|\mathbf{v}(k_j)\|} \), and the edge weight is a combination of positional and semantic factors: \( \mu_{ij} = \theta_{ij} + \sigma_{ij} \), with \( \sigma_{ij} = \gamma \cdot S(k_i, k_j) \).
We further refine these relationships by adjusting edge weights contextually using related words and apply positional decay \( \kappa_{ij} = e^{-\lambda l_i} \) to emphasize proximity. The final edge weight \( \omega_{ij} = \tau_{ij} \cdot \kappa_{ij} \) integrates positional, semantic, and contextual factors.

Once the graph is constructed, we compute a ranking score for each word based on its centrality using a weighted PageRank-inspired algorithm~\cite{bianchini2005inside,mihalcea2004textrank}. The initial score for each word \( k_i \) is \( r_0(k_i) = (1 - \alpha_{\text{page}}) \), where \( \alpha_{\text{page}} \) is a damping factor. The score of each word is influenced by its neighbors, with the contribution from a neighboring word \( k_j \) calculated as \( r_{c}(k_i, k_j) = \frac{\omega_{ij} \cdot r(k_j)}{\sum_{k_m \in \mathcal{N}(k_j)} \omega_{jm}} \), where \( \omega_{ij} \) is the edge weight, \( r(k_j) \) is the neighbor's score, and \( \mathcal{N}(k_j) \) represents the neighbors of \( k_j \). The score for each word \( k_i \) is updated iteratively using \( r(k_i) = (1 - \alpha_{\text{page}}) + \alpha_{\text{page}} \cdot \sum_{k_j \in \mathcal{N}(k_i)} r_{c}(k_i, k_j) \), with \( \alpha_{\text{page}} \) controlling the influence of neighbors. This process continues until convergence, where all words are ranked by their final scores \( r(k_i) \). The top-$K$ keywords, corresponding to the most significant regions in the image, are then selected.

\noindent{\textbf{Region Mapping and Perturbation. }} The top-ranked keywords guide a modified Segment Anything Model (SAM)~\cite{kirillov2023segment}, mapping each keyword \( k_i \) to its corresponding visual regions. We then apply diffusion noise to perturb these regions~\cite{ho2020denoising}. For each region \( R_i \) of the image \( m \), the perturbation follows the diffusion process \( m' = \sqrt{\alpha_{\text{diff}, t}} \cdot R_i + \sqrt{1 - \alpha_{\text{diff}, t}} \cdot \epsilon \), where \( \alpha_{\text{diff}, t} = \prod_{j=0}^{t} \beta_{\text{diff}, j} \) (with \( \beta_{\text{diff}, t} \in (0, 1) \) as a hyperparameter controlling the noise schedule), and \( \epsilon \) is random noise sampled from a standard normal distribution. This perturbation selectively distorts key regions, encouraging the model to focus on critical areas. For images whose keywords fail to locate any visual region, we add diffusion noise across all image areas to generate rejected image data. These perturbed images \( m' \) serve as rejected preference images in DPO, refining the model's alignment with fine-grained visual details.

\begin{table*}[!ht]
\centering
\resizebox{0.8\textwidth}{!}{%
\begin{tabular}{lcc|cc|cccc}
\toprule
\multirow{2}{*}{Methods} & \multicolumn{2}{c|}{\textbf{MMHalBench}} & \multicolumn{2}{c|}{\textbf{Object HalBench}} & \multicolumn{4}{c}{\textbf{AMBER}} \\ 
\cline{2-3} \cline{4-5} \cline{6-9} 
& \textbf{Score $\uparrow$} & \textbf{HalRate $\downarrow$} & \textbf{CHAIR$_s$ $\downarrow$} & \textbf{CHAIR$_i$ $\downarrow$} & \textbf{CHAIR$_s$ $\downarrow$} & \textbf{Cover. $\uparrow$} & \textbf{HalRate $\downarrow$} & \textbf{Cog. $\downarrow$} \\
\midrule
\rowcolor[HTML]{E5E1E0}\multicolumn{9}{l}{\textit{\textbf{7B MLLMs}}} \\ 
\midrule
LLaVA-v1.6-7B~\cite{liu2024visual}  & 2.46 & 0.52 & 16.4 & 9.4 & 9.1 & \textbf{61.7} & 50.2 & 4.7 \\
LLaVA-v1.5-7B~\cite{liu2024visual} & 2.07 & 0.59 & 53.6 & 25.2 & 7.8 & 51.0 & 36.4 & 4.2 \\
+ HACL~\cite{jiang2024hallucination} & 2.13 & 0.50 & -- & -- & -- & -- & -- & -- \\
+ POVID ~\cite{zhou2024aligning} & 2.08 & 0.56 & 48.1 & 24.4 & -- & -- & -- & -- \\
+ OPERA~\cite{huang2024opera} & 2.15 & 0.54 & 45.1 & 22.3 & -- & -- & -- & -- \\
+ VCD~\cite{leng2024mitigating} & 2.04 & 0.58 & 48.0 & 22.3 & -- & -- & -- & -- \\
+ DPO~\cite{rafailov2024direct} & 2.14 & 0.65 & 49.0 & 13.0 & 6.5 & 55.5 & 34.5 & 2.3 \\
+ mDPO~\cite{wang2024mdpo} & 2.39 & 0.54 & 35.7 & 9.8 & 4.4 & 52.4 & 24.5 & 2.4 \\
+ EOS~\cite{yue2024less} & 2.03 & 0.59 & 40.3 & 17.8 & 5.1 & 49.1 & 22.7 & 2.0 \\
+ HA-DPO~\cite{zhao2023beyond} & 1.97 & 0.59 & 39.9 & 19.9 & 6.7 & 49.8 & 30.9 & 3.3 \\
+ HALVA~\cite{sarkar2024mitigating} & 2.08 & 0.60 & 46.6 & 53.0 & 6.6 & 53.0 & 33.2 & 3.4 \\
\rowcolor[HTML]{aad1cd} LLaVA-v1.5-7B + MFPO & \underline{2.69} {\textcolor[HTML]{990000}{\scriptsize($\uparrow$0.62)}} & \underline{0.49} {\textcolor[HTML]{990000}{\scriptsize($\downarrow$0.1)}} & \underline{13.4} {\textcolor[HTML]{990000}{\scriptsize($\downarrow$40.2)}} & \underline{6.6} {\textcolor[HTML]{990000}{\scriptsize($\downarrow$18.6)}} & \underline{4.1} {\textcolor[HTML]{990000}{\scriptsize($\downarrow$3.7)}} & 55.7 {\textcolor[HTML]{990000}{\scriptsize($\uparrow$4.7)}} & \underline{22.5} {\textcolor[HTML]{990000}{\scriptsize($\downarrow$13.9)}} & \underline{1.9} {\textcolor[HTML]{990000}{\scriptsize($\downarrow$2.3)}} \\
\rowcolor[HTML]{aad1cd} LLaVA-v1.6-7B + MFPO & \textbf{2.89} {\textcolor[HTML]{990000}{\scriptsize($\uparrow$0.43)}} & \textbf{0.45} {\textcolor[HTML]{990000}{\scriptsize($\downarrow$0.07)}} & \textbf{10.6} {\textcolor[HTML]{990000}{\scriptsize($\downarrow$5.8)}} & \textbf{5.1} {\textcolor[HTML]{990000}{\scriptsize($\downarrow$4.3)}} & \textbf{3.1} {\textcolor[HTML]{990000}{\scriptsize($\downarrow$6.0)}} & \underline{58.8} {\textcolor[HTML]{990000}{\scriptsize($\downarrow$2.9)}} & \textbf{18.7} {\textcolor[HTML]{990000}{\scriptsize($\downarrow$31.5)}} & \textbf{1.1} {\textcolor[HTML]{990000}{\scriptsize($\downarrow$3.6)}} \\
\midrule
\rowcolor[HTML]{E5E1E0}\multicolumn{9}{l}{\textit{\textbf{>=13B MLLMs}}} \\ 
\midrule
GPT-4V~\cite{achiam2023gpt} & 3.49 & 0.28 & 13.6 & {7.3} & 4.6 & 67.1 & 30.7 & 2.6 \\
MiniGemini-34B~\cite{li2024mini} & 3.08 & 0.38 & 14.5 & 8.0 & -- & -- & -- & -- \\
\hdashline
Qwen-VL-Chat~\cite{bai2023qwen} & \underline{2.89} & \underline{0.43} & 36.0 & 21.3 & 6.6 & \underline{53.2} & 31.0 & 2.9 \\
LLaVA-v1.5-13B~\cite{liu2024improved} & 2.42 & 0.53 & 46.3 & 22.6 & 7.8 & 51.0 & 36.4 & 4.2 \\
+ RLHF-V~\cite{yu2024rlhf} & 2.81 & 0.49 & \underline{12.2} & \underline{7.5} & \underline{6.3} & 46.1 & \underline{25.1} & \underline{2.1} \\
% + HSA-DPO~\cite{xiao2024detecting} & 2.61 & 0.48 & \textbf{5.2} & \textbf{3.2} & \textbf{2.1} & 47.3 & \textbf{13.4} & \textbf{1.2} \\
+ HALVA~\cite{sarkar2024mitigating} & 2.84 & 0.48 & -- & -- & 6.4 & 52.6 & 30.4 & 3.2 \\
\hdashline
\rowcolor[HTML]{aad1cd} LLaVA-v1.5-13B + MFPO & \textbf{2.94} {\textcolor[HTML]{990000}{\scriptsize($\uparrow$0.52)}} & \textbf{0.42} {\textcolor[HTML]{990000}{\scriptsize($\downarrow$0.11)}} & \textbf{11.4} {\textcolor[HTML]{990000}{\scriptsize($\downarrow$34.9)}} & \textbf{4.6} {\textcolor[HTML]{990000}{\scriptsize($\downarrow$18.0)}} & \textbf{3.4} {\textcolor[HTML]{990000}{\scriptsize($\downarrow$4.4)}} & \textbf{56.1} {\textcolor[HTML]{990000}{\scriptsize($\uparrow$5.1)}} & \textbf{19.4} {\textcolor[HTML]{990000}{\scriptsize($\downarrow$17.0)}} & \textbf{1.4} {\textcolor[HTML]{990000}{\scriptsize($\downarrow$2.8)}} \\
\bottomrule
\end{tabular}}
\caption{Results for MMHalBench, Object HalBench, and AMBER benchmarks. \textcolor[HTML]{990000}{ Text in red} highlights comparisons before and after incorporating MFPO.}
\label{tab:main_results}
\end{table*}

\begin{table*}[htbp!]
\centering
\resizebox{0.8\textwidth}{!}{%
\begin{tabular}{l|ccccccccc}
\toprule
 & overall & attribute & adversarial & comparison & counting & relation & environment & holistic & other \\
\midrule
LLaVA-RLHF-7B & 2.05  & 2.92 & 1.83 & 2.42 & 1.92 & 2.25 & 2.25& 1.75 & 1.08 \\
LLaVA-RLHF-13B & \underline{2.53} & \underline{3.33} & \underline{2.67} & 1.75 & {2.25} & \underline{2.33} & \underline{3.25} & \textbf{2.25} & \textbf{2.42} \\
\hdashline
LLaVA-v1.5-7B & 2.07 & 3.08 & 1.08 & \textbf{2.58} & \underline{2.25} & {2.0} & 3.0 & 1.42 & 1.33 \\
\rowcolor[HTML]{aad1cd}+ MFPO & \textbf{2.69} & \textbf{3.33} & \textbf{3.67} & \underline{2.42} & \textbf{2.25} & \textbf{2.75} & \textbf{3.42} & \underline{2.00} & \underline{1.83} \\
\bottomrule
\end{tabular}
}
\caption{Performance comparison across different dimensions in MMHalBench.}
\label{tab:performance_comparison}
\end{table*}

\subsection{MFPO Training Process}

We leverage the generated image preference data for a unified optimization of the policy \( \pi_\theta \) across both text and image.

\noindent\textbf{Text Preference Alignment.} 
The DPO policy objective for text preference optimization is:
\begin{equation}
\resizebox{0.5\textwidth}{!}{$
\begin{aligned}
    L_{\text{text}}(\pi_\theta; \pi_{\text{ref}}) = - \mathbb{E}_{D} \left[ \log \sigma \left( \beta \log \frac{\pi_\theta(y_w|t, m)}{\pi_{\text{ref}}(y_w|t, m)} - \beta \log \frac{\pi_\theta(y_l|t, m)}{\pi_{\text{ref}}(y_l|t, m)} \right) \right],
\end{aligned}
$}
\end{equation}
where \( t \) and \( m \) represents the original text and image inputs. \( y_w \) and \( y_l \) are the preferred and less preferred outputs. \( \pi_{\text{ref}} \) is the reference policy.

\noindent\textbf{Image Preference Alignment.} For image preference optimization, we use the original image \( m \) and the perturbed image \( m' \) generated in earlier steps. The image preference loss is defined as:
\begin{equation}
\resizebox{0.5\textwidth}{!}{$
\begin{aligned}
    L_{\text{image}}(\pi_\theta; \pi_{\text{ref}}) = - \mathbb{E}_{D} \left[ \log \sigma \left( \beta \log \frac{\pi_\theta(y_w|t, m)}{\pi_{\text{ref}}(y_w|t, m)} - \beta \log \frac{\pi_\theta(y_w|t, m')}{\pi_{\text{ref}}(y_w|t, m')} \right) \right].
\end{aligned}
$}
\end{equation}
Note that we only use the preferred response \( y_w \) for image preference alignment.

\noindent\textbf{Margin Loss for Stability.} To address potential instability during the joint optimization of text and image preferences, we introduce a margin loss to penalize situations where both chosen and rejected responses experience a reduction in reward, drawing inspiration from previous research ~\cite{meng2024simpo,chen2024noise,wang2024mdpo}.
The margin loss is defined as:
\begin{equation}
    L_{\text{margin}} = - \log \sigma \left( \beta \log \frac{\pi_\theta(y_w|t,m)}{\pi_{\text{ref}}(y_w|t,m)} - \eta \right),
\end{equation}
where \( \eta \) is the margin parameter that enforces a greater separation between positive (chosen) and negative (rejected) responses. The total loss for the training phase is defined as \( L_{\text{total}} = L_{\text{text}} + L_{\text{image}} + L_{\text{margin}} \),  ensuring that both text and image preferences are jointly optimized while maintaining stability in their alignment.
\subsection{Easy-to-Hard Iterative Alignment}
We introduce an Easy-to-Hard Iterative Alignment algorithm to stabilize the training of MFPO.

\noindent{\textbf{Entropy Calculation.}} Leveraging semantic entropy~\cite{venhuizen2019semantic,farquhar2024detecting}, we estimate the uncertainty of the model's responses. Given a probability distribution of predicted answers \( P = \{p_1, p_2, \dots, p_n\} \), where \( p_i \) represents the probability of the \( i \)-th predicted token, the entropy \( H \) is calculated as \( H(P) = - \sum_{i=1}^{n} p_i \log(p_i) \). This measure quantifies the uncertainty inherent in the model.

\noindent{\textbf{Sorting by Difficulty.}}
After calculating entropy for all training samples, we rank the training dataset according to their entropy scores, where higher values denoting more challenging inputs. We then divide the dataset into three distinct difficulty levels: "easy", "medium", and "hard".

\noindent{\textbf{Iterative Alignment.}}
By progressively moving from easy to hard examples, the feedback distribution is iteratively updated. This ensures that the model first learns simpler patterns to build a foundation before tackling more complex cases. As the training progresses, the alignment becomes more refined, allowing the model to effectively handle harder examples.

\section{Experiements}
\subsection{Experimental Setup}

\noindent\textbf{Implementation Details.} We use LLaVA-v1.5 as the backbone for all experiments and include the latest top-performing LLaVA-v1.6 to validate the effectiveness of our method. The training consists of three stages: the first two stages follow standard LLaVA training, while MFPO is introduced in the third stage. Here, we construct image preference data based on Section \ref{gen_da}, using text preference data from RLHF-V~\cite{yu2024rlhf}, and apply MFPO optimization. Details are in Supplementary Section 4.

\noindent\textbf{Evaluation.} We evaluate the trustworthiness and general model capabilities with comprehensive experiments. For trustworthiness, we employ three widely used benchmarks to evaluate trustworthiness reflecting the degree of hallucination. Object HalBench~\cite{rohrbach2018object} was used to assess common object hallucinations in image descriptions, with CHAIR scores for both response-level (CHAIRs) and object-level (CHAIRi) hallucination rates. MMHal-Bench~\cite{sun2023aligning} measured response quality and hallucination rates by comparing model outputs with human responses and object labels. AMBER~\cite{wang2023llm} evaluated generative tasks, providing metrics on CHAIR variants, object coverage, and cognitive hallucination rates. For general capabilities, we employ the LLaVA-Bench~\cite{liu2024visual} for systematic comprehension, encompassing two categories: conversation, detailed description. See details in Supplementary Section 5. 

\noindent{\textbf{Baselines.}} We compare our model with state-of-the-art baselines under different settings. For general MLLMs, we include LLaVA1.5~\cite{liu2024improved}, Qwen-VL-Chat~\cite{bai2023qwen}, LLaVA1.6, and MiniGemini~\cite{li2024mini}. We also compare with preference-feedback models such as POVID~\cite{zhou2024aligning}, RLHF-V~\cite{yu2024rlhf}, Silkie~\cite{sun2023aligning} and DPO~\cite{rafailov2024direct}, as well as feedback-independent methods such as VCD~\cite{leng2024mitigating}. We also benchmark against GPT-4V~\cite{achiam2023gpt}. See details in Supplementary Section 6.

\subsection{Main Results}
\noindent\textbf{Comparison to Existing Methods.} 
The main results are presented in Table~\ref{tab:main_results}. With MFPO integration, both LLaVA-v1.5 and LLaVA-v1.6 show consistent improvements across all three trustworthiness evaluation datasets. On MMHalBench and Object HalBench, our approach achieves state-of-the-art performance among all 7B-parameter models. As shown in Table~\ref{tab:performance_comparison}, MFPO enhances performance across all metrics, demonstrating universal improvements in MLLM trustworthiness. These results highlight MFPO's ability to balance preference optimization across text and image modalities, leading to superior trustworthiness.

Notably, LLaVA-v1.5-7B+MFPO achieves trustworthiness levels comparable to or exceeding those of larger models, such as LLaVA-v1.5-13B and MiniGemini-34B. Furthermore, while baseline LLaVA-v1.5-7B and LLaVA-v1.5-13B underperform GPT-4V across all eight metrics, MFPO integration enables both models to outperform GPT-4V on five out of eight metrics. This underscores MFPO's capability to significantly enhance the trustworthiness of smaller MLLMs, bridging the performance gap with much larger models.

\begin{table}[t]
\centering
\small
\resizebox{0.4\textwidth}{!}{%
\begin{tabular}{l|cc}
\toprule
Method & Conversation & Captioning   \\
\midrule
LLaVA-1.5 & {53.3} & 53.4 \\
+ VIfeedback & 51.3 & 49.3  \\
+ Human-Preference & 49.6 & 43.3  \\
+ RLHF-V & \underline{55.8} & \underline{56.1}  \\
\midrule
\rowcolor[HTML]{aad1cd}Ours & \textbf{65.8} & \textbf{60.0} \\
\bottomrule
\end{tabular}}
\caption{Comparison of different methods in LLaVA-Bench.}
\label{tab:method_comparison}
\end{table}

\noindent\textbf{General Perception Evaluation.} To verify whether MFPO can enhance general perception capability of MLLM, we analyze different preference collection strategies on LLaVA-Bench benchmark. We focus on the performance related to conversation and detailed descriptions, which represent general perception abilities. As shown in Table \ref{tab:method_comparison}, our method outperforms both LLaVA-v1.5 and other alignment methods, indicating that our approach not only enhances trustworthiness but also improves overall perception capability of the model.

\subsection{Ablation Study and Analysis}
\begin{table}[t]
    \centering
    \resizebox{0.4\textwidth}{!}{
    \begin{tabular}{c|c|c|c|c|c|c}
        \toprule
        $L_{\text{text}}$ & $L_{\text{image}}$ & $L_{\text{margin}}$ & \multicolumn{2}{c|}{\textbf{MMHalBench}} & \multicolumn{2}{c}{\textbf{Object HalBench}} \\
        \cline{4-7}
        & & & \textbf{Score} $\uparrow$ & \textbf{HalRate}$\downarrow$ & \textbf{CHAIR}$_s$ $\downarrow$ & \textbf{CHAIR}$_i$ $\downarrow$ \\
        \midrule
        \checkmark & x & \checkmark & 2.46 & 0.53 & {21.9} & 10.9  \\
        x & \checkmark & \checkmark & 2.34 & 0.56 & 24.4 & 13.1  \\
        \checkmark & \checkmark & x & \underline{2.61} & \underline{0.50} & \underline{16.9} & \underline{7.9}  \\
        \checkmark & \checkmark & \checkmark & \textbf{2.69} & \textbf{0.49} & \textbf{13.4} & \textbf{6.6}  \\
        \bottomrule
    \end{tabular}
    }
    \caption{Ablation study of different loss compositions.}
    \label{tab:loss_compositions}
\end{table}

\begin{table}[t!]
\centering
\resizebox{0.4\textwidth}{!}{
\begin{tabular}{c|cc|cc}
\toprule
Margin & \multicolumn{2}{c|}{\textbf{MMHalBench}} & \multicolumn{2}{c}{\textbf{Object HalBench}} \\
& \textbf{Score} $\uparrow$ & \textbf{HalRate} $\downarrow$ & \textbf{CHAIR}$_s$ $\downarrow$ & \textbf{CHAIR}$_i$ $\downarrow$ \\
\midrule
0 (Ours) & \textbf{2.69} & \textbf{0.49} & \textbf{13.4} & \textbf{6.6} \\
0.2 & 2.58 & \underline{0.50} & \underline{15.7} & \underline{7.1} \\
0.4 & \underline{2.63} & 0.52 & 16.0 & 7.9 \\
\bottomrule
\end{tabular}
}
\caption{Comparison of different margins.}
\label{tab:margin}
\end{table}
\noindent\textbf{Ablation on Joint Text-Visual Reward Loss.} We conduct experiments to validate the loss composition in MFPO. As shown in Table~\ref{tab:loss_compositions}, optimizing only the text modality ($L_{\text{text}}$) improves performance but overfits to text preferences, while optimizing only the image modality ($L_{\text{image}}$) addresses the text-image preference imbalance but underperforms due to lack of text optimization. This highlights the necessity of joint text-visual optimization in multimodal tasks.

\begin{figure}[t!]
    \centering
    \includegraphics[width=1\linewidth]{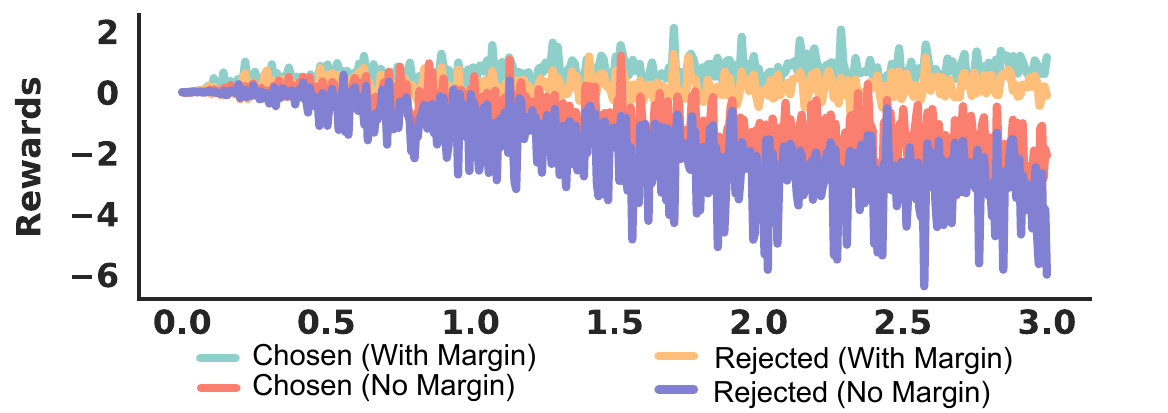}
    \caption{Rewards with and without margin loss in 3 epochs.}
    \label{fig:reward_com}
\end{figure}

\noindent\textbf{Ablation on Margin Loss.} Adding margin loss ($L_{\text{margin}}$) further improves performance. Figure~\ref{fig:reward_com} shows that without $L_{\text{margin}}$, rewards for both chosen and rejected responses decrease simultaneously, while $L_{\text{margin}}$ stabilizes rewards by penalizing reductions for chosen responses. Table~\ref{tab:margin} demonstrates consistent performance gains across margin values (0, 0.2, 0.4), with the best results at 0. 

\begin{table}[ht!]
    \small
    \centering
    \renewcommand{\arraystretch}{0.6} % 减少行间距
    \resizebox{0.8\columnwidth}{!}{% Adjust width as needed
    \begin{tabular}{ccccc}
        \toprule
        {} & \multicolumn{2}{c}{MMHalbench} & \multicolumn{2}{c}{Object Halbench} \\
        \cmidrule(lr){2-3} \cmidrule(lr){4-5}
        Ratio & Score$\uparrow$ & HalRate$\downarrow$ & CHAIRs$\downarrow$ & CHAIRi$\downarrow$\\
        \midrule
        1:1:1 & \textbf{2.69} & \textbf{0.49} & \textbf{13.4} & \textbf{6.6} \\
        1:5:1 & 2.66 & 0.50 & 13.9 & 6.9 \\
        5:1:1 & 2.59 & 0.55 & 15.6 & 8.2 \\
        1:1:5 & 2.62 & 0.52 & 14.5 & 7.0 \\
        \bottomrule
    \end{tabular}
    }
    \caption{Ablation of Ratio \(L_{\text{text}}:L_{\text{image}}:L_{\text{margin}}\)}
    \label{tab:ablation_ratio} 
\end{table}
\noindent\textbf{Ablation on Loss Ratio.} In Table~\ref{tab:ablation_ratio}, we present an ablation study on the weighting of different loss components. A balanced ratio of 1:1:1 yields the best results.

\begin{table}[h]
\centering
\resizebox{0.4\textwidth}{!}{
\begin{tabular}{c|cc|cc}
\toprule
Training Method & \multicolumn{2}{c|}{\textbf{MMHalBench}} & \multicolumn{2}{c}{\textbf{Object HalBench}} \\
& \textbf{Score} $\uparrow$ & \textbf{HalRate} $\downarrow$ & \textbf{CHAIR}$_s$ $\downarrow$ & \textbf{CHAIR}$_i$ $\downarrow$ \\
\midrule
Easy-to-hard & \textbf{2.69} & \textbf{0.49} & \textbf{13.4} & \textbf{6.6} \\  
End-to-end & \underline{2.53} & \underline{0.53} & \underline{16.0} & \underline{8.0} \\
\bottomrule
\end{tabular}
}
\caption{Comparison of different training methods.}
\label{tab:training_methods}
\end{table}

\begin{figure}[t!]
    \centering
    \includegraphics[width=1\linewidth]{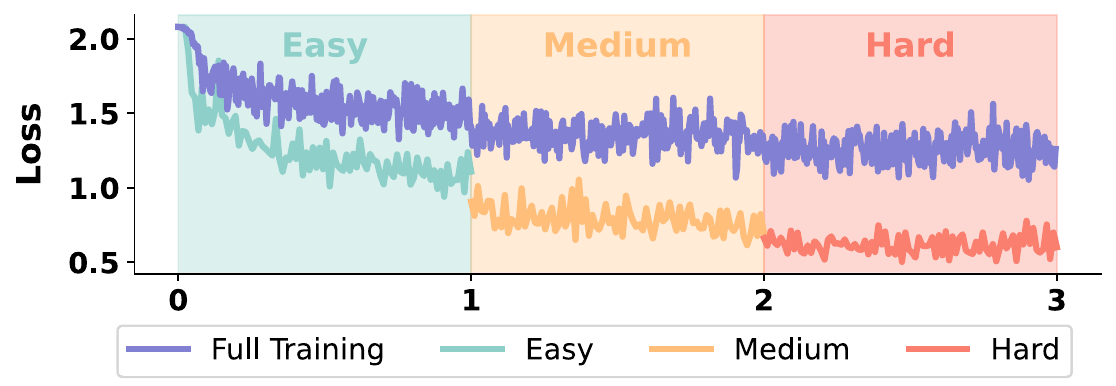}
    \caption{Loss comparison in 3 epochs.}
    \label{fig:loss_com}
\end{figure}

\noindent\textbf{Ablation of Easy-to-Hard Alignment Scheme.} 
We evaluate the effectiveness of the easy-to-hard alignment scheme against simultaneous training on all data. Table~\ref{tab:training_methods} shows that our approach consistently outperforms the traditional method. Figure~\ref{fig:loss_com} further reveals that the easy-to-hard strategy achieves better convergence, with losses steadily decreasing across easy, medium, and hard phases, indicating more efficient alignment learning. The loss reduction is fastest in the easy phase and gradually slows in later phases. Moreover, the easy-to-hard training curve exhibits greater stability compared to full training, demonstrating its stability and efficiency.

\begin{figure}[!t]
    \centering
    \includegraphics[width=1\linewidth]{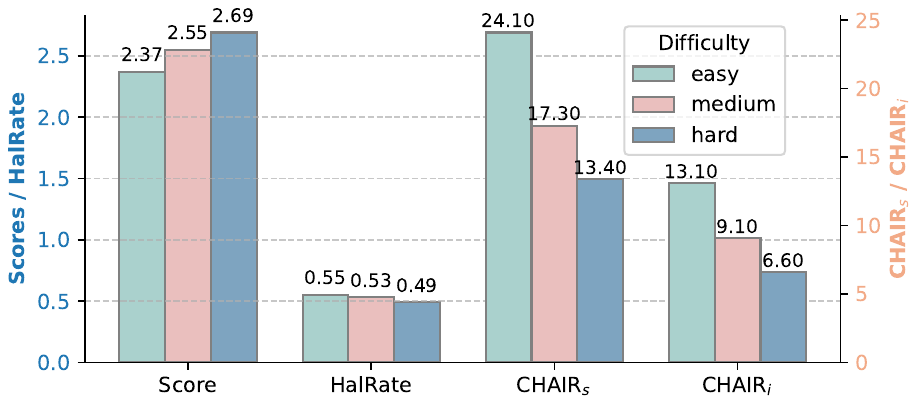}
    \caption{Each stage's performance.}
    \label{tab:stage_perfor}
\end{figure}

\begin{table}[t!]
\centering
\resizebox{0.45\textwidth}{!}{
\begin{tabular}{c|cc|cc}
\toprule
Construction Method & \multicolumn{2}{c|}{\textbf{MMHalBench}} & \multicolumn{2}{c}{\textbf{Object HalBench}} \\
& \textbf{Score} $\uparrow$ & \textbf{HalRate} $\downarrow$ & \textbf{CHAIR}$_s$ $\downarrow$ & \textbf{CHAIR}$_i$ $\downarrow$ \\
\midrule
Global & \underline{2.56} & 0.56 & 21.7 & 10.3 \\
Random 20\% & 2.45 & \underline{0.52} & \underline{16.5} & \underline{8.5} \\
Ours & \textbf{2.69} & \textbf{0.49} & \textbf{13.4} & \textbf{6.6} \\
\bottomrule
\end{tabular}
}
\caption{Comparison of different image preference data construction.}
\label{tab:img_cons}
\end{table}

\noindent\textbf{How Does Noise Proportion Affect Preference Optimization Balance?} 
We study how noise granularity and proportion affect preference optimization balance by comparing three methods: applying global diffusion noise to the entire image, adding noise to 20\% of random image regions, and our fine-grained noise addition strategy. Table~\ref{tab:img_cons} demonstrates that image preference data construction enhances image-based optimization during alignment. However, as noise granularity becomes coarser, optimization increasingly favors text preferences, leading to performance degradation. Coarse-grained methods oversimplify image feature optimization, impairing image-text alignment. In contrast, fine-grained construction shifts optimization toward the image modality, achieving a more balanced multimodal alignment.

\noindent\textbf{How Does Keyword Selection Perform?} 
To evaluate the accuracy of extracting key visual keyphrases from the original text preference data using a multipartite graph, we randomly sample 100 examples and manually verify their correctness. We find that 86\% of the extracted keyphrases correspond to important objects in the associated preference data. In the remaining 14\% of cases, the selected keywords are not the most critical objects; however, using SAM to add diffusion noise to these regions still generates finely dispreferred images. We attribute the effectiveness of the multipartite graph to the relatively short text length and simple object relationships in the preference datasets, which enable robust keyword selection.

\noindent\textbf{How Does SAM Perform on Region Mapping?} 
To validate SAM's accuracy in region mapping, we randomly sample 100 images for manual evaluation. SAM achieves over 90\% accuracy, with the remaining 10\% of failures occurring when keywords refer to objects requiring contextual understanding beyond SAM's scope. In such cases, we apply global diffusion noise to the entire image. While this approach is suboptimal, the resulting dispreferred images still improve model performance. Furthermore, we experiment with more advanced segmentation models, such as LISA~\cite{Lai_2024_CVPR}, which achieves over 95\% accuracy in region mapping. Models with reasoning-based segmentation capabilities offer greater robustness to diverse keywords, enabling the generation of more refined dispreferred images.

\begin{figure}[t!]
    \centering
    \includegraphics[width=1\linewidth]{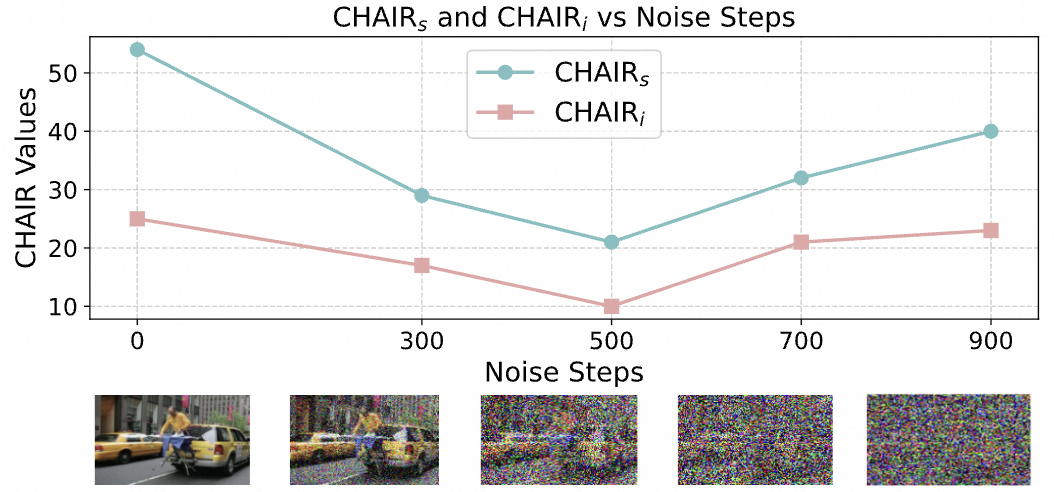}
    \caption{Performance of LLaVA-v1.5 under different noise levels.}
    \label{fig:loss_noise}
\end{figure}

\noindent\textbf{How Do Different Noise Levels Affect Performance?} 
We investigate the impact of varying noise levels on preference optimization. As shown in Figure~\ref{fig:loss_noise}, adding diffusion noise to the entire image improves performance, with the best results achieved at 500 steps. At this level, the image undergoes an appropriate level of corruption: it avoids excessive noise that would erase critical details and make dispreferred images too easily distinguishable, while also preventing insufficient noise that would result in minimal image changes, failing to establish meaningful rewards for image optimization. 

\begin{table}[t!]
    \centering
    \renewcommand{\arraystretch}{1} % 减少行间距
    \resizebox{0.9\columnwidth}{!}{% Adjust width as needed
    \begin{tabular}{ccccc}
        \toprule
        {} & \multicolumn{2}{c}{MMHalbench} & \multicolumn{2}{c}{Object Halbench} \\
        \cmidrule(lr){2-3} \cmidrule(lr){4-5}
        Methods & Score$\uparrow$ & HalRate$\downarrow$ & CHAIRs$\downarrow$ & CHAIRi$\downarrow$\\
        \midrule
        mDPO (RLHF-V)& 2.42 & 0.56 & 29.3 & 9.2 \\
        \rowcolor[HTML]{aad1cd}MFPO (RLHF-V)& \textbf{2.69} & 0.49 & \textbf{13.4} & \textbf{6.6} \\
        mDPO (Silkie) & 2.39 & 0.54 & 35.7 & 9.8 \\
        \rowcolor[HTML]{aad1cd}MFPO (Silkie)& 2.67 &\textbf{0.46} & 14.7 & 7.1\\
        \bottomrule
    \end{tabular}
    }
    \caption{Further Comparision of mDPO and MFPO.}
    \label{tab:further}
\end{table}

\noindent\textbf{How to Achieve Modality-Fair Optimization?}
We investigate modality-fair optimization from both data and loss perspectives. On the data side, we analyze the impact of noisy region proportion and noise intensity in images. Our results show that region-level noise is more effective than global noise in guiding preference toward the visual modality. Moreover, moderate noise levels yield a better optimization balance than extremely high or low noise. On the loss side, jointly optimizing text and image rewards facilitates balanced modality alignment while preserving overall performance.

\noindent\textbf{More Comparisions to mDPO.} 
While mDPO demonstrated its effectiveness by only training on Silkie, Table~\ref{tab:further} reports MFPO’s performance when trained on Silkie and mDPO’s performance when trained on RLHF-V. This ensures a fair comparison under equivalent dataset conditions. The results show that {dataset differences have minimal impact on MFPO's performance}, and {MFPO consistently outperforms mDPO across both datasets}, further validating MFPO's effectiveness.

\section{Conclusion}
In this paper, we investigate the limitation of text preference optimization and propose Modality-Fair Preference Optimization (MFPO) for trustworhty aligntment in MLLMs. We constructed a new dataset with fine-grained image preference, and implemented balanced preference optimization between text and image using a multi-stage alignment strategy alongside novel loss functions. Extensive experiments and analysis demonstrate that MFPO notebly diminishes hallucinations and attains state-of-the-art performance across trustworthiness and general capability benchmarks.

\section*{Acknowledgments}
This work is supported by the National Natural Science Foundation of China (Grant No. 12326612, 62476241), the Natural Science Foundation of Zhejiang Province, China (Grant No. LZ23F020008), and the Zhejiang University-Angelalign Inc. R\&D Center for Intelligent Healthcare.

\bibliographystyle{named}
\bibliography{ijcai25}

\end{document}